\documentclass[runningheads]{eccv2020/llncs}
\usepackage{graphicx}
\usepackage{eccv2020/comment}
\usepackage{amsmath,amssymb} 
\usepackage{amsmath}
\usepackage{color}
\usepackage{booktabs}
\usepackage{array}
\usepackage{subcaption}
\usepackage[pagebackref=true,breaklinks=true,letterpaper=true,colorlinks,bookmarks=false,citecolor=blue,linkcolor=blue]{hyperref}

\usepackage[ruled,lined]{algorithm2e}

\newcommand{\fancyname}{SimAug}

\newcommand{\context}{\mathcal{H}}
\newcommand{\real}{\mathbb{R}}
\DeclareMathOperator*{\argmax}{argmax}

\newcolumntype{C}[1]{>{\centering\arraybackslash}p{#1}}

\begin{document}
\pagestyle{headings}
\mainmatter
\def\ECCVSubNumber{1850}  

\title{\textit{\fancyname}: Learning Robust Representations from Simulation for Trajectory Prediction} 

\titlerunning{\textit{\fancyname}}
%
\author{Junwei Liang\inst{1}\orcidID{0000-0003-2219-5569} \and
Lu Jiang\inst{2} \and
Alexander Hauptmann\inst{1}}
\authorrunning{Junwei Liang, Lu Jiang, Alexander Hauptmann}
%

\institute{$^1$Carnegie Mellon University\hspace{20pt}$^2$Google Research \\
\email{\{junweil,alex\}@cs.cmu.edu, lujiang@google.com}}
\maketitle


\begin{figure}[ht]
	\centering
		\includegraphics[width=\hsize]{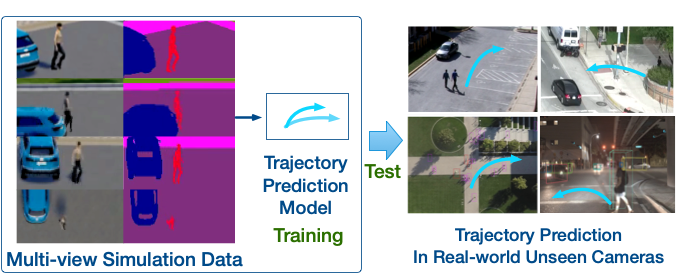}
	\caption{Illustration of pedestrian trajectory prediction in unseen cameras. We propose to learn robust representations only from 3D simulation data that could generalize to real-world videos captured by unseen cameras.}
	\label{fig:intro}
\end{figure}

\begin{abstract}

This paper studies the problem of predicting future trajectories of people in unseen cameras of novel scenarios and views.
We approach this problem through the real-data-free setting in which the model is trained only on 3D simulation data and applied out-of-the-box to a wide variety of real cameras.
We propose a novel approach to learn robust representation through augmenting the simulation training data such that the representation can better generalize to unseen real-world test data. The key idea is to mix the feature of the hardest camera view with the adversarial feature of the original view.
We refer to our method as \textit{\fancyname}. 
We show that \textit{\fancyname} achieves promising results on three real-world benchmarks using zero real training data, and state-of-the-art performance in the Stanford Drone and the VIRAT/ActEV dataset when using in-domain training data.
Code and models are released at
\url{https://next.cs.cmu.edu/simaug}.

\keywords{Trajectory Prediction, 3D Simulation, Robust Learning, Data Augmentation, Representation Learning, Adversarial Learning}
\end{abstract}

\section{Introduction}
Future trajectory prediction~\cite{kitani2012activity,alahi2016social,gupta2018social,liang2019peeking,sadeghian2018sophie,lee2017desire,liang2020garden} is a fundamental problem in video analytics, which aims at forecasting a pedestrian's future path in the video in the next few seconds. Recent advancements in future trajectory prediction have been successful in a variety of vision applications such as
self-driving vehicles~\cite{bansal2018chauffeurnet,chai2019multipath,chang2019argoverse}, safety monitoring~\cite{liang2019peeking}, robotic planning~\cite{rhinehart2017first,rhinehart2018r2p2}, among others.

A notable bottleneck for existing works is that the current model is closely coupled with the video cameras on which it is trained, and generalizes poorly on new cameras with novel views or scenes. 
For example, prior works have proposed various models to forecast a pedestrian's trajectories in video cameras of different types such as stationary outdoor cameras~\cite{oh2011large,liang2019focal,alahi2016social,gupta2018social,lerner2007crowds,luber2010people}, drone cameras~\cite{sadeghian2018sophie,deo2020trajectory,li2019way}, ground-level egocentric cameras~\cite{yagi2018future,rhinehart2017first,styles2019multiple}, or dash cameras~\cite{mangalam2019disentangling,styles2019forecasting,chang2019argoverse}. 
However, existing models are all separately trained and tested within one or two datasets, and there have been no attempts at successfully generalizing the model across datasets of novel camera views.
This bottleneck significantly hinders the application whenever there is a new camera because it requires annotating new data to fine-tune the model, resulting in a procedure that is not only expensive but also tardy in deploying the model.

An ideal model should be able to disentangle human behavioral dynamics from specific camera views, positions, and scenes. It should produce robust trajectory prediction despite the variances in these camera settings. 
Motivated by this idea, in this work, we learn a robust representation for future trajectory prediction that can generalize to unseen video cameras. Different from existing works, we study a \emph{real-data-free} setting where a model is trained only on synthetic data but tested, out of the box, on unseen real-world videos, without further re-training or fine-tuning the model. 
Following the success of learning from simulation~\cite{ruiz2018learning,de2017procedural,varol2019synthetic,zhang2019rsa,gaidon2016virtual,richter2016playing}, our synthetic data is generalized from a 3D simulator, called CARLA~\cite{dosovitskiy2017carla}, which anchors to the static scene and dynamic elements in the VIRAT/ActEV videos~\cite{oh2011large}. By virtue of the 3D simulator, we can generate multiple views and pixel-precise semantic segmentation labels for each training trajectory, 
as illustrated in Figure~\ref{fig:intro}.
Meanwhile, following the previous works~\cite{sadeghian2018sophie,liang2020garden}, scene semantic segmentation is used instead of RGB pixels to alleviate the influence of different lighting conditions, scene textures, subtle noises produced by camera sensors, etc. At test time, we extract scene features from real videos using pretrained segmentation models.
The use of segmentation features is helpful but is insufficient for learning robust representation for future trajectory prediction.

To tackle this issue, we propose a novel data augmentation method called \emph{SimAug} to augment the features of the simulation data with the goal of learning robust representation to various semantic scenes and camera views in real videos. 
To be specific, first, after representing each training trajectory by high-level scene semantic segmentation features, we defend our model from adversarial examples generated by white-box attack methods~\cite{goodfellow2014explaining}. 
Second, to overcome the changes in camera views, we generate multiple views for the same trajectory, and encourage the model to focus on overcoming the ``hardest'' view to which the model has learned. Following~\cite{jiang2018mentornet,jiang2019synthetic}, the classification loss is adopted and the view with the highest loss is favored during training. 
Finally, the augmented trajectory is computed as a convex combination of the trajectories generated in previous steps.
Our trajectory prediction backbone model is built on a recent work called Multiverse~\cite{liang2020garden}. The final model is trained to minimize the empirical vicinal risk over the distribution of augmented trajectories.
Our method is partially inspired by recent robust deep learning methods using adversarial training~\cite{kurakin2016adversarial,cheng2020advaug}, Mixup~\cite{zhang2017mixup}, and MentorMix~\cite{jiang2019synthetic}.

We empirically validate our model, which is trained only on simulation data, 
on three real-world benchmarks for future trajectory prediction: VIRAT/ActEV \cite{oh2011large,2018trecvidawad}, Stanford Drone~\cite{robicquet2016learning}, and Argoverse~\cite{chang2019argoverse}. 
These benchmarks represent three distinct camera views: 45-degree view, top-down view and dashboard camera view with ego-motions.
The results show our method performs favorably against baseline methods including standard data augmentation, adversarial learning, and imitation learning. 
Notably, our method achieves better results compared to the state-of-the-art on the VIRAT/ActEV and Stanford Drone benchmark. 
Our code and models are released at \url{https://next.cs.cmu.edu/simaug}. To summarize, our contribution is threefold:
\begin{itemize}
    \item We study a new setting of future trajectory prediction in which the model is trained only on synthetic data and tested, out of the box, on any unseen real video with novel views or scenes.
    \item We propose a novel and effective approach to augment the representation of trajectory prediction models using multi-view simulation data.
    \item Ours is the first work on future trajectory prediction to demonstrate the efficacy of training on 3D simulation data, and establishes new state-of-the-art results on three public benchmarks.
\end{itemize}

\section{Related Work}
\noindent\textbf{Trajectory prediction.}
Recently there is a large body of work on predicting person future trajectories in a variety of scenarios.
Many works~\cite{alahi2016social,xue2018ss,zhang2019sr,liang2019peeking,liang2020garden,sadeghian2018sophie} focused on modeling person motions in videos recorded with stationary cameras.
Datasets like VIRAT/ActEV~\cite{oh2011large}, ETH/UCY~\cite{lerner2007crowds,luber2010people} and Stanford Drone~\cite{robicquet2016learning} have been used for evaluating pedestrian trajectory prediction.
For example, Social-LSTM~\cite{alahi2016social} added social pooling to model nearby pedestrian trajectory patterns. 
Social-GAN~\cite{gupta2018social} added an adversarial network~\cite{goodfellow2014generative} on Social-LSTM to generate diverse future trajectories. 
Several works focused on learning the effects of the physical scene, e.g., people tend to walk on the sidewalk instead of grass. 
Kitani et al.~\cite{kitani2012activity} used Inverse Reinforcement Learning to forecast human trajectory. 
SoPhie~\cite{sadeghian2018sophie} combined deep neural network features from scene semantic segmentation model and generative adversarial network (GAN) using attention to model person trajectory.
More recent works~\cite{kooij2014context,yagi2018future,ma2017forecasting,liang2019peeking} have attempted to predict person paths by utilizing individuals' visual features instead of considering them as points in the scene.
For example, Liang et al.~\cite{liang2020garden} proposed to use abstract scene semantic segmentation features for better generalization.
Meanwhile, many works~\cite{lee2017desire,sadeghian2018car,bansal2018chauffeurnet,hong2019rules,zhao2019multi,makansi2019overcoming,li2019way,rhinehart2018r2p2} have been proposed for top-down view videos for trajectory prediction.
Notably, the Stanford Drone Dataset (SDD)~\cite{robicquet2016learning} is used in many works~\cite{sadeghian2018sophie,deo2020trajectory,li2019way} for trajectory prediction with drone videos.
Other works have also looked into pedestrian prediction in dashcam videos~\cite{mangalam2019disentangling,styles2019forecasting,kooij2014context,lee2017desire} and first-person videos~\cite{yagi2018future,styles2019multiple}.
Many vehicle trajectory datasets~\cite{caesar2019nuscenes,chang2019argoverse,yu2018bdd100k} have been proposed as a result of self-driving's surging popularity.

\noindent\textbf{Learning from 3D simulation data.}
As the increasing research focus in 3D computer vision~\cite{zhang2015fast,liang2017event,shah2018airsim,dosovitskiy2017carla,richter2016playing,ros2016synthia,heess2017emergence}, many research works have used 3D simulation for training and evaluating real-world tasks~\cite{gaidon2016virtual,de2017procedural,wu2019revisiting,zhu2017target,sun2019stochastic,liang2020garden,sun2018multi,bak2018domain,kar2019meta,chen2019data}.
Many works~\cite{qiu2017unrealcv,gaidon2016virtual,de2017procedural} were proposed to use data generated from 3D simulation for video object detection, tracking, and action recognition analysis. 
For example, Sun et al.~\cite{sun2019stochastic} proposed a forecasting model by using a gaming simulator.
AirSim ~\cite{shah2018airsim} and CARLA ~\cite{dosovitskiy2017carla} were proposed for robotic autonomous controls for drones and vehicles.
Zeng et al.~\cite{zeng2019adversarial} proposed to use 3D simulation for adversarial attacks. RSA~\cite{zhang2019rsa} used randomized simulation data for human action recognition.
The ForkingPaths dataset~\cite{liang2020garden} was proposed for evaluating multi-future trajectory prediction. Human annotators were asked to control agents in a 3D simulator to create a multi-future trajectory dataset. 

\noindent\textbf{Robust Deep Learning.}
Traditional domain adaptation approaches \cite{bousmalis2017unsupervised,ganin2016domain,tzeng2017adversarial,kang2019contrastive} may not be applicable as our target domain is considered ``unseen'' during training. Methods for learning using privileged information~\cite{lambert2018deep,vapnik2015learning,lopez2015unifying,luo2018graph} is not applicable for a similar reason. Closest to ours is robust deep learning methods. In particular, our approach is inspired by the following methods: (i) \textit{adversarial training}~\cite{goodfellow2014explaining,madry2017towards,xie2019feature,zeng2019adversarial} to defend the adversarial attacks generated on-the-fly during training using gradient-based methods~\cite{madry2017towards,goodfellow2014explaining,tramer2017ensemble,cheng2019robust}; (ii) data augmentation methods to overcome unknown variances between training and test examples such as Mixup~\cite{zhang2017mixup}, MentorMix~\cite{jiang2019synthetic}, AugMix~\cite{cheng2020advaug}, etc; (iii) example re-weighting or selection~\cite{jiang2018mentornet,ren2018learning,jiang2015self,liang2016learning,northcutt2019confident} to mitigate network memorization.
Different from prior work, ours uses 3D simulation data as a new perspective for data augmentation and is carefully designed for future trajectory prediction.

\begin{figure}[ht]
	\centering
		\includegraphics[width=\textwidth]{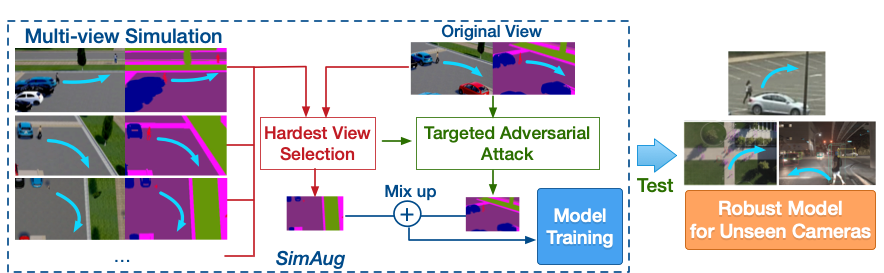}
	\caption{Overview of our method \emph{SimAug} that is trained on simulation and tested on real unseen videos. Each training trajectory is represented by multi-view segmentation features extracted from the simulator. \emph{SimAug} mixes the feature of the hardest camera view with the adversarial feature of the original view.}
	\label{fig:method}
\end{figure}

\begin{figure}[]
	\centering
		\includegraphics[width=\textwidth]{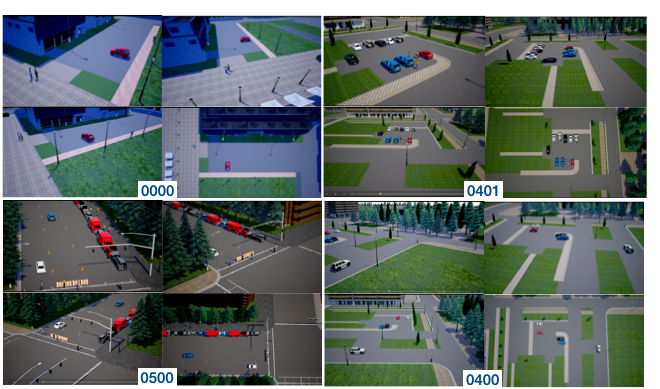}
	\caption{Visualization of the multi-view 3D simulation data used in \textit{\fancyname} training. Data generation process is described in Section~\ref{sec:data}. We use 4 camera views from 4 scenes defined in~\cite{liang2020garden}. ``0400'' and ``0401'' scene have overlapping views. The top-left views are the original views from VIRAT/ActEV dataset.}
	\label{fig:data}
\end{figure}

\section{Approach}
\label{sec:approach}
In this section, we describe our approach to learn robust representation for future trajectory prediction, which we call \textit{\fancyname}. Our goal is to train a model only on simulation training data that can effectively predict the future trajectory in the real-world test videos that are unseen during training.

\subsection{Problem Formulation}
We focus on predicting the locations of a single agent for multiple steps into the future. Given a sequence of historical video frames $V_{1:h}$ of the past $h$ steps and the past agent locations $L_{1:h}$ in training, we learn a probabilistic model on simulation data to estimate $P(L_{h+1:T}|L_{1:h}, V_{1:h})$ for $T-h$ steps into the future. At test time, our model takes as input an agent's observable past $(V_{1:h}, L_{1:h})$ in real videos to predict the agent's future locations $L_{h+1:T} = \{y_{h+1}, \ldots, y_T\}$, where $y_t$ is the location coordinates. As the test real videos are unseen during training, the model is supposed to be invariant to the variances in semantic scenes, camera views, and camera motions.

\subsection{Training Data Generation From Simulation}
\label{sec:data}
Our model is trained only on simulation data. To ensure high-level realism, the training trajectories are generated by CARLA~\cite{dosovitskiy2017carla}, an open-source 3D simulator built on top of the state-of-the-art game engine \textit{Unreal Engine 4}. We use the trajectories from the Forking Paths dataset~\cite{liang2020garden} that are semi-manually recreated from the VIRAT/ActEV benchmark that projects real-world annotations to the 3D simulation world. Note that it is not our intention to build an exact replica of the real-world scene, nor it is necessary to help train a model for real-world tasks as suggested in previous works~\cite{gaidon2016virtual,ros2016synthia,liang2020garden,zhang2019rsa}.

With CARLA, we record multiple views of the same trajectory of different camera angles and positions. For a trajectory $(V_{1:T}, L_{1:T})$ in original view, let $\mathcal{S} = \{(V_{1:T}^{(i)}, L_{1:T}^{(i)})\}_{i=1}^{|\mathcal{S}|}$ denote a set of additional views for the same trajectory.
In our experiments, we use four camera parameters pre-specified in~\cite{liang2020garden}, including three 45-degree views and one top-down view. 
We use a total of 4 scenes shown in Fig.~\ref{fig:data}. 
The ground-truth location varies under different camera views i.e. $L_{1:T}^{(i)} \ne L_{1:T}^{(j)}$ for $i \ne j$.
Note that these camera positions and angles are defined in~\cite{liang2020garden} specifically for VIRAT/ActEV dataset.
The top-down view cameras in Stanford Drone dataset~\cite{robicquet2016learning} are still considered unseen to the model since the scenes and camera positions are quite different.

In simulation, we also collect the ground-truth scene semantic segmentation for $K=13$ classes including sidewalk, road, vehicle, pedestrian, etc. At test time, we extract the semantic segmentation feature from real videos using a pre-trained model with the same number of class labels per pixel. To be specific, we use the Deeplab model \cite{chen2017deeplab} trained on the ADE20k \cite{zhou2017scene} dataset and keep its weights frozen. To bridge the gap between real and simulated video frames, we represent all trajectory $V_{1:T}$ as a sequence of scene semantic segmentation features, following previous works~\cite{liang2019peeking,liang2020garden,deo2020trajectory,sadeghian2018sophie}. As we show in our experiments, the use of segmentation features is helpful but is still insufficient for learning the robust representation.

\subsection{Multi-view Simulation Augmentation (\textit{\fancyname})}
\label{sec:simaug}
In this subsection, we first describe \textit{\fancyname} for learning robust representations. Our trajectory prediction backbone model is built on the Multiverse model~\cite{liang2020garden} and will be discussed in Section~\ref{sec:multiverse}.

Given a trajectory in its original view $(V_{1:T}, L_{1:T})$, we generate a set of additional views in $\mathcal{S} = \{(V_{1:T}^{(i)}, L_{1:T}^{(i)})\}_{i=1}^{|\mathcal{S}|}$ as described in the previous section, where $V_{t}^{(i)}$ represents the scene semantic feature of view $i$ at time $t$. $L_{1:T}^{(i)}$ is a sequence of ground-truth locations for the $i$-th view. 

Each time given a camera view, we use it as an anchor to search for the ``hardest'' view that is most inconsistent with what the model has learned. 
Inspired by~\cite{jiang2018mentornet}, we use the classification loss as the criteria and compute:
\begin{align}
    \label{eqn:select_c}
    j^* = \argmax_{j \in [1,|\mathcal{S}|]} \{ 
    \mathcal{L}_\text{cls}(V_{1:h} + \delta, L_{h+1:T}^{(j)}) \},
\end{align}
where $\delta$ is the $\ell_{\infty}$-bounded random perturbation applied to the input features.  $\mathcal{L}_\text{cls}$ is the location classification loss used in our backbone Multiverse model and will be discussed in the next subsection.

Then for the original view, we generate an adversarial trajectory by the targeted-FGSM attack~\cite{kurakin2016adversarial}:
\begin{align}
    V_{1:h}^{adv} &= V_{1:h} - \epsilon  \cdot \text{sign}(\nabla_{V_{1:h}} \mathcal{L}_\text{cls}( V_{1:h} + \delta, L_{h+1:T}^{(j^*)})),
    \label{eqn:adv}
\end{align}
where $\epsilon$ is the hyper-parameter. 
The attack tries to make the model predict the future locations in the selected ``hardest'' camera view rather than the original view.
In essence, the resulting adversarial feature is ``warped'' to the ``hardest'' camera view by a small perturbation.
By defending against such adversarial trajectory, our model learns representations that are robust against variances in camera views.  

Finally, we mix up the trajectory locations of the selected view and the adversarial trajectory locations by a convex combination function~\cite{zhang2017mixup} over their features and one-hot location labels.
\begin{align}
\begin{split}
    V_{1:h}^{aug} &= \lambda \cdot V_{1:h}^{adv} + (1-\lambda) \cdot V_{1:h}^{(j^*)} \\
    y_t^{aug} &= \lambda \cdot \text{one-hot}(y_t) + (1-\lambda) \cdot\text{one-hot}(y_t^{(j^*)}) \quad \forall t \in [h+1,T] \\
    L_{h+1:T}^{aug} &= [y_{h+1}^{aug}, \ldots, y_T^{aug}]
    \label{eqn:mixup}
\end{split}
\end{align}
where $[y_{h+1}, \cdots, y_T]=L_{h+1:T}$ are the ground-truth locations of the original view. The $\text{one-hot}(\cdot)$ function projects the location in $xy$ coordinates into an one-hot embedding over the predefined grid used in our backbone trajectory prediction model. Please find the details in~\cite{liang2020garden}.
Following~\cite{zhang2017mixup}, $\lambda$ is drawn from a Beta distribution $\text{Beta}(\alpha, \alpha)$ controlled by the hyper-parameter $\alpha$.

The algorithm for training with one training step is listed in Algorithm~\ref{alg:simaug}.
To train robust models to various camera views and semantic scenes, we learn representations over augmented training trajectories to overcome (i) feature perturbations (Step 3 and 5) (ii) targeted adversarial attack (Step 5), and (iii) the ``hardest'' feature from other views (Step 4).
By the mix-up operation in Eq.~\eqref{eqn:mixup}, 
our model is trained to minimize the empirical vicinal risk over a new distribution constituted by the generated augmented trajectories, which is proved to be useful in improving model robustness to real-world distributions under various settings~\cite{jiang2019synthetic}.

\begin{algorithm}
\SetKwData{Left}{left}\SetKwData{This}{this}\SetKwData{Up}{up}
\SetKwInOut{Input}{Input}\SetKwInOut{Output}{Output}
\LinesNumbered
\Input{Mini-batch of trajectories; hyper-parameters $\alpha$ and $\epsilon$}
\Output{Classification loss $\mathcal{L}_\text{cls}$ computed over augmented trajectories}
\BlankLine
\For{each trajectory $(V_{1:T}, L_{1:T})$ in the mini-batch}{
    Generate trajectories from additional views $\mathcal{S}= \{(V_{1:T}^{(i)}, L_{1:T}^{(i)})\}$\;
    Compute the loss for each camera view $\mathcal{L}_\text{cls}( V_{1:h} + \delta, L_{h+1:T}^{(j)})$\;
    Select the view with the largest loss $j^*$ by Eq.~\eqref{eqn:select_c} \;
    Generate an adversarial trajectory $V_{1:h}^{adv}$ by Eq.~\eqref{eqn:adv}\;
    Mix up ($V_{1:h}^{adv}$, $L_{h+1:T}$) and ($V_{1:h}^{(j^*)}$, $L_{h+1:T}^{(j*)}$) by Eq.~\eqref{eqn:mixup}\;
    Compute $\mathcal{L}_\text{cls}$ over the augmented trajectory ($V^{aug}_{1:h}$, $L_{h+1:T}^{aug}$) from Step 6\;
}
\Return  averaged $\mathcal{L}_\text{cls}$ over the augmented trajectories
\caption{\small Multi-view Simulation Adversarial Augmentation (\textit{\fancyname})}
\label{alg:simaug}
\end{algorithm}

\subsection{Backbone Model for Trajectory Prediction}
\label{sec:multiverse}

We employ Multiverse~\cite{liang2020garden} as our backbone network, a state-of-the-art multi-future trajectory prediction model. Although we showcase the use of \emph{\fancyname} to improve the robustness of Multiverse, \emph{\fancyname} is a general approach that can be applied to other trajectory prediction models.

\noindent\textbf{Input Features.}
The model is given the past locations, $L_{1:h}$, and the scene, $V_{1:h}$. 
Each ground-truth location $L_t$ is encoded by an one-hot vector $y_{t} \in \mathbb{R}^{HW}$ representing the nearest cell in a 2D grid of size $H \times W$.
In our experiment, we use a grid scale of $36 \times 18$.
Each video frame $V_t$ is encoded as semantic segmentation feature of size $H \times W \times K$ where $K=13$ is the total number of class labels as in~\cite{liang2020garden,liang2019peeking}.
As discussed in the previous section, we use \textit{\fancyname} to generate augmented trajectories $(V_{1:h}^{aug}, L_{1:h}^{aug})$ as our training features.

\noindent\textbf{History Encoder.} 
A convolutional RNN~\cite{xingjian2015convolutional,wang2019eidetic} is used to get the final spatial-temporal feature state $H_t \in \mathbb{R}^{H \times W \times d_{enc}}$, where $d_{enc}$ is the hidden size.
The context is a concatenation of the last hidden state and the historical video frames, $\context=[H_h,V_{1:h}]$.

\noindent\textbf{Location Decoder.}
After getting the context $\context$, a coarse location decoder is used to predict locations at the level of grid cells at each time-instant by:
\begin{align}
    \hat{y}_{t} = \text{softmax}(f_c(\context, H^c_{t-1})) \in \mathbb{R}^{HW}
    \label{eqn:fc}
\end{align}
where $f_c$ is the convolutional recurrent neural network (ConvRNN) with graph attention proposed in~\cite{liang2020garden} and $H^c_t$ is the hidden state of the ConvRNN.
Then a fine location decoder is used to predict a continuous offset in $\real^2$, which specifies a ``delta''
from the center of each grid cell, to get a fine-grained location prediction:
\begin{align}
    \hat{O}_t = \text{MLP}(f_o(\context, H^o_{t-1})) \in \mathbb{R}^{HW \times 2},
    \label{eqn:fo}
\end{align}
where $f_o$ is a separate ConvRNN and $H^o_t$ is its hidden state.
To compute the final prediction location, we use
\begin{align}
    \hat{L}_t = Q_{g} + \hat{O}_{tg}
    \label{eq:L_t}
\end{align}
where $g=\argmax \hat{y}_{t}$ is the index of the selected grid cell, $Q_{g} \in \real^2$ is the center of that cell,
and $\hat{O}_{tg} \in \real^2$ is the predicted offset for that cell at time $t$.

\noindent\textbf{Training.} 
We use \textit{\fancyname} (see Section~\ref{sec:simaug}) to generate $L_{h+1:T}^{aug}= \{y_{h+1}^{aug}, \ldots, y_T^{aug} \}$ as labels for training.
For the coarse decoder, the cross-entropy loss is used:
\begin{equation}
    \mathcal{L}_\text{cls} = -\frac{1}{T} \sum_{t=h+1}^{T} 
    \sum^{HW}_{c=1} y^{aug}_{tc} \log(\hat{y}_{tc})
    \label{eqn:loss_cls}
\end{equation}
For the fine decoder, we use the original ground-truth location label $L_{h+1:T}$:
\begin{equation}
 \mathcal{L}_\text{reg} = \frac{1}{T} \sum_{t=h+1}^{T} 
 \sum^{HW}_{c=1}
 \text{smooth}_{l_1}(O_{tc}, \hat{O}_{tc})
\end{equation}
where 
$O_{tc} = L_t - Q_{c}$ 
is the delta between the ground true location and the center of the $c$-th grid cell. 
The final loss is then calculated using
\begin{align}
\mathcal{L}(\theta) = \mathcal{L}_\text{cls} + \lambda_1 \mathcal{L}_\text{reg} + \lambda_2 \|\theta\|_2^2
\end{align}
where $\lambda_2$ controls the $\ell_2$ regularization (weight decay),
and $\lambda_1=0.5$ is used
to balance the regression and classification losses.

\section{Experiments}
\label{sec:exp}

In this section, we evaluate various methods, including our \textit{\fancyname} method, on three public video benchmarks of real-world videos captured under different camera views and scenes: the VIRAT/ActEV~\cite{2018trecvidawad,oh2011large} dataset, the Stanford Drone Dataset (SDD)~\cite{robicquet2016learning}, and the autonomous driving dataset Argoverse~\cite{chang2019argoverse}.
We demonstrate the efficacy of our method for unseen cameras in Section~\ref{sec:exp_main} and how our method can also improve state-of-the-art when fine-tuned on the real training data in Section~\ref{sec:exp_sdd} and Section~\ref{sec:exp_actev}.

\subsection{Evaluation Metrics}
\label{sec:metrics}
Following prior works~\cite{alahi2016social,liang2020garden}, we utilize two common metrics for trajectory prediction evaluation. Let $L^{i}=L^{i}_{t=(h+1)\cdots T}$ be the true future trajectory for the $i^{th}$ test sample, and $\hat{L}^{ik}$ be the corresponding $k^{th}$ prediction sample, for $k \in [1,K]$.

\noindent i) \textit{Minimum Average Displacement Error Given K Predictions} (minADE\textsubscript{K}): for each true trajectory sample $i$,
we select the closest $K$ predictions,
and then measure its average error:
\begin{equation}
    \text{minADE}_K = \frac{  \sum^{N}_{i=1} \min_{k=1}^K \sum^{T}_{t=h+1} \lVert L^i_t - \hat{L}^{ik}_t \rVert_{2}}{N \times (T-h)}
\end{equation}

\noindent ii) \textit{Minimum Final Displacement Error Given K Predictions} (minFDE\textsubscript{K}): similar to minADE\textsubscript{K}, but we only consider the predicted points and the ground truth point at the final prediction time instant:
\begin{equation}
    \text{minFDE}_K = \frac{  \sum^{N}_{i=1} \min_{k=1}^K \lVert L^{i}_{T} -  \hat{L}^{ik}_{T} \rVert_{2}}{N}
\end{equation}
\noindent iii) \textit{Grid Prediction Accuracy} (Grid\_Acc):
As our base model also predicts coarse grid locations as described in Section~\ref{sec:multiverse}, we also evaluate the accuracy between the predicted grid $\hat{y}_t$ and the ground truth grid $y_t$. This is an intermediate metric and hence is less indicative than the minADE\textsubscript{K} and minFDE\textsubscript{K}.

\subsection{Main Results}
\label{sec:exp_main}
\noindent\textbf{Dataset \& Setups.} 
We compare \textit{\fancyname} with classical data augmentation methods as well as adversarial learning methods to train robust representations.
All methods are trained using the same backbone on the same \textit{simulation training data} described in Section~\ref{sec:data}, and tested on the same benchmarks. Real videos are not allowed to be used during training except in our finetuning experiments. For VIRAT/ActEV and SDD, we use the standard test split as in \cite{liang2019peeking,liang2020garden} and \cite{sadeghian2018sophie,deo2020trajectory}, respectively.
For Argoverse, we use the official validation set from the 3D tracking task, and the videos from the ``ring\_front\_center'' camera are used.

These datasets have different levels of difficulties. VIRAT/ActEV is the easiest one because its training trajectories have been projected in the simulation training data. SDD is more difficult as its camera positions and scenes are different from the training data. Argoverse is the most challenging one with distinct scenes, camera views, and ego-motions.


Following the setting in previous works~\cite{liang2019peeking,alahi2016social,gupta2018social,alahi2016social,gupta2018social,sadeghian2018sophie,makansi2019overcoming,liang2020garden,deo2020trajectory}, 
the models observe 3.2 seconds (8 frames) of every pedestrian and predict the future 4.8 seconds (12 frames) of the person trajectory. 
We use the pixel values for the trajectory coordinates as it is done in~\cite{yagi2018future,liang2019peeking,lee2017desire,chai2019multipath,li2019way,makansi2019overcoming,bansal2018chauffeurnet,hong2019rules,deo2020trajectory}. 
By default, we evaluate the top $K=1$ future trajectory prediction of all models.

\noindent\textbf{Baseline methods.} 
We compare \textit{\fancyname} with the following baseline methods for learning robust representations. All methods are built on the same backbone network and trained using the same simulation training data.
\textit{Base Model} is the trajectory prediction model proposed in~\cite{liang2020garden}.
\textit{Standard Aug} is the base model trained with standard data augmentation techniques including horizontal flipping and random input jittering. 
\textit{Fast Gradient Sign Method (FGSM)} is the base model trained with adversarial examples generated by the targeted-FGSM attack method~\cite{goodfellow2014explaining}. Random labels are used for the targeted-FGSM attack. 
\textit{Projected Gradient Descent (PGD)} is learned with an iterative adversarial learning method~\cite{madry2017towards,xie2019feature}. The number of iterations is set to 10 and other hyper-parameters following~\cite{xie2019feature}.

\noindent\textbf{Implementation Details.}
We use $\alpha=0.2$ for the $\text{Beta}$ distribution in Eq~\eqref{eqn:mixup} and we use $\epsilon=\delta=0.1$ in Eq~\eqref{eqn:adv}.
As the random perturbation is small, we do not normalize the perturbed features and the normalized features yield comparable results.
All models are trained using Adadelta optimizer~\cite{zeiler2012adadelta} with an initial learning rate of 0.3 and a weight decay of 0.001.
Other hyper-parameters for the baselines are the same as the ones in~\cite{liang2020garden}.

\noindent\textbf{Quantitative Results.}
Table~\ref{table:main} shows the evaluation results. 
Our method performs favorably against other baseline methods across all evaluation metrics on all three benchmarks. 
In particular, ``Standard Aug'' seems to be not generalizing well to unseen cameras.
FGSM improves significantly on the ``Grid\_Acc'' metric but fails to translate the improvement to final location predictions. 
\textit{\fancyname} is able to improve the model overall stemming from the effective use of multi-view simulation data.
All other methods are unable to improve trajectory prediction on Argoverse, whose data characteristics include ego-motions and distinct dashboard-view cameras.
The results substantiate the efficacy of \textit{\fancyname} for future trajectory prediction in unseen cameras. Note as the baseline methods use the same features as ours, the results indicate the use of segmentation features is insufficient for learning robust representations.

\begin{figure}[ht]
	\centering
		\includegraphics[width=\textwidth]{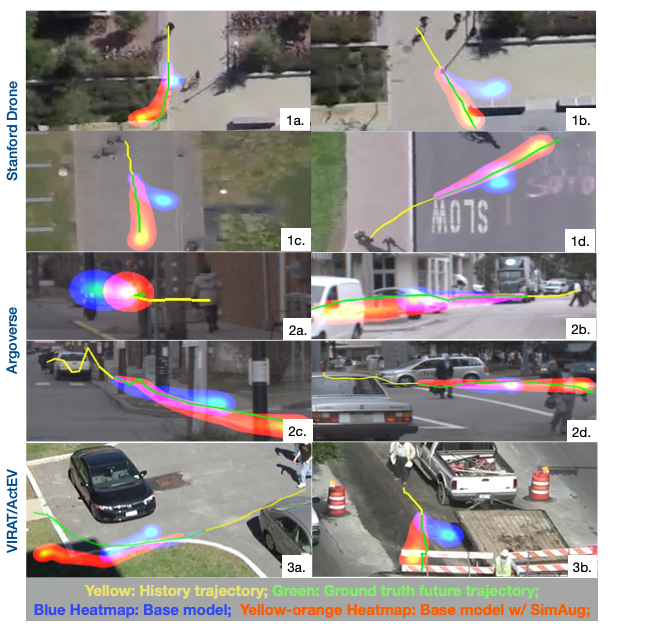}
	\caption{Qualitative analysis. Trajectory predictions from different models are colored and overlaid in the same image. See text for details. }
	\label{fig:qualitative}
\end{figure}

\noindent\textbf{Qualitative Analysis.}
We visualize outputs of the base model with and without \textit{\fancyname} in Fig.~\ref{fig:qualitative}. We show visualizations on all three datasets. 
In each image, the yellow trajectories denote historical trajectories and the green ones are ground truth future trajectories. Outputs of the base model without {\fancyname} are colored with blue heatmaps and the yellow-orange heatmaps are from the same model with {\fancyname}.
As we see, the base model with {\fancyname} augmentation yields more accurate trajectories for turnings (Fig.~\ref{fig:qualitative} 1a., 3a.) while without it the model sometimes predicts the wrong turns (Fig.~\ref{fig:qualitative} 1b., 1c., 2a., 3a., 3b.).
In addition, the length of {\fancyname} predictions is more accurate (Fig.~\ref{fig:qualitative} 1d., 2b., 2c., 2d.).

\begin{table}[]
\centering
\caption{Comparison to the standard data augmentation method and recent adversarial learning methods on three datasets. We report three metrics: Grid\_Acc($\uparrow$)/minADE\textsubscript{1}($\downarrow$)/minFDE\textsubscript{1}($\downarrow$). The units of ADE/FDE are pixels. All methods are built on the same backbone model in~\cite{liang2020garden} and trained using the same multi-view simulation data described in Section~\ref{sec:data}.}
\setlength\tabcolsep{2mm}
\begin{tabular}{lccc}
\toprule
Method & VIRAT/ActEV & Stanford Drone  & Argoverse    \\ 
\midrule
Base Model~\cite{liang2020garden} & 44.2\%/26.2/49.7 & 31.4\%/21.9/42.8  &  26.6\%/69.1/183.9 \\ 
Standard Aug &   45.5\%/25.8/48.3 & 21.3\%/23.7/47.6  &  28.9\%/70.9/183.4 \\ 
PGD~\cite{madry2017towards,xie2019feature} &  47.5\%/25.1/48.4 & 28.5\%/21.0/42.2 & 25.9\%/72.8/184.0 \\
FGSM~\cite{goodfellow2014explaining} &  48.6\%/25.4/49.3 & 42.3\%/19.3/39.9 & 29.2\%/71.1/185.4\\
SimAug &  \textbf{51.1}\%/\textbf{21.7}/\textbf{42.2} &  \textbf{45.4\%}/\textbf{15.7}/\textbf{30.2} & \textbf{30.9\%}/\textbf{67.9}/\textbf{175.6}\\
\bottomrule
\end{tabular}
\label{table:main}
\end{table}

\subsection{State-of-the-Art Comparison on Stanford Drone Dataset}
\label{sec:exp_sdd}
In this section, we compare our \textit{\fancyname} model with the state-of-the-art generative models, including Social-LSTM~\cite{alahi2016social}, Social-GAN~\cite{gupta2018social}, DESIRE~\cite{lee2017desire}, and SoPhie~\cite{sadeghian2018sophie}. We also compare with the imitation learning model, IDL~\cite{li2019way}, and the inverse reinforcement learning model, P2T\textsubscript{IRL}~\cite{deo2020trajectory} for trajectory prediction on the Stanford Drone Dataset. Following previous works, we evaluate the minimal errors over $K=20$ predictions.

\noindent\textbf{Results \& Analysis.} The results are shown in Table~\ref{table:sdd_actev} (a), where \textit{\fancyname} is built on top of the \textit{Multiverse} model. 
As it shows, \textit{\fancyname} model trained only on the simulation data (second to the last row) achieves comparable or even better performance than other state-of-the-art models that are trained on in-domain real videos. By further fine-tuning on the learned representations of \textit{\fancyname}, we achieve the state-of-the-art performance on the Stanford Drone Dataset. 
The promising results demonstrate the efficacy of \textit{\fancyname} for future trajectory prediction in unseen cameras.

\begin{table}[]
\centering
\caption{State-of-the-art comparison on the Stanford Drone Dataset (SDD) and on the VIRAT/ActEV dataset. Numbers are minimal errors over 20 predictions for SDD and minimal errors over the top prediction for VIRAT/ActEV. Baseline models are all trained on real videos and their numbers are taken from \cite{sadeghian2018sophie,deo2020trajectory}. ``SimAug'' is trained only using simulation data and ``SimAug*'' is further finetuned on the real training videos. 
``Multiverse'' in the bottom rows is trained only with simulation data.}
\begin{subtable}[t]{.5\textwidth} 
\centering
\begin{tabular}[t]{lcc}
\toprule
Method & \scriptsize{minADE\textsubscript{20}($\downarrow$)}   & \scriptsize{minFDE\textsubscript{20}} ($\downarrow$)      \\
\midrule
Social-LSTM~\cite{alahi2016social} & 31.19 & 56.97 \\
Social-GAN~\cite{gupta2018social} & 27.25   &  41.44 \\ 
DESIRE~\cite{lee2017desire} &  19.25 & 34.05 \\ 
SoPhie~\cite{sadeghian2018sophie} &  16.27 & 29.38 \\ 
Multiverse~\cite{liang2020garden} & 14.78 & 27.09 \\
IDL~\cite{li2019way} &  13.93 & 24.40 \\ 
P2T\textsubscript{IRL}~\cite{deo2020trajectory} &  12.58 & 22.07 \\ 
\midrule
SimAug & 12.03  & 23.98 \\
SimAug* & \textbf{10.27}  & \textbf{19.71} \\
\bottomrule
\end{tabular}
\caption{Stanford Drone Dataset}
\end{subtable}
\begin{subtable}[t]{.48\textwidth}
\centering
\begin{tabular}[t]{lcc}
\toprule
Method & \scriptsize{minADE\textsubscript{1}($\downarrow$)}   & \scriptsize{minFDE\textsubscript{1} ($\downarrow$)}      \\ 
\midrule
Social-LSTM~\cite{alahi2016social}  & 23.10 & 44.27 \\
Social-GAN~\cite{gupta2018social}  & 30.42   & 60.70  \\ 
Next~\cite{liang2019peeking}  & 19.78 & 42.43\\
Multiverse~\cite{liang2020garden} & 18.51 & 35.84 \\
\midrule
Multiverse~\cite{liang2020garden} & 22.94 & 43.35 \\
SimAug & 21.73  & 42.22 \\
SimAug*  & \textbf{17.96}  & \textbf{34.68} \\
\bottomrule
\end{tabular}
\caption{VIRAT/ActEV}
\end{subtable}
\label{table:sdd_actev}
\end{table}

\subsection{State-of-the-Art Comparison on VIRAT/ActEV}
\label{sec:exp_actev}
In this section, we compare our \textit{\fancyname} model with state-of-the-art models on VIRAT/ActEV.
Following the previous work~\cite{liang2020garden}, we compute the errors for the top $K=1$ prediction. 
Experimental results are shown in Table~\ref{table:sdd_actev} (b), where all models in the top four rows are trained on the real-world training videos in VIRAT/ActEV. 
Our model trained on simulation data achieves competitive performance and outperforms \textit{Multiverse}~\cite{liang2020garden} model that is trained on the same data.
With fine-tuning, which means using exactly the same training data without any extra annotation of real trajectories compared to ~\cite{alahi2016social,gupta2018social,liang2019peeking,liang2020garden}, we achieve the best performance on the VIRAT/ActEV benchmark.



\subsection{Ablation Experiments}

We test various ablations of our approach to validate our design decisions.
Results are shown in Table~\ref{table:ablation}, where the top-1 prediction is used in the evaluation.
We verify four key design choices by removing each, at a time, from the full model.
The results show that by introducing viewpoint selection (Eq.~\eqref{eqn:select_c}) and adversarial perturbation (Eq.~\eqref{eqn:adv}), our method improves model generalization. 

(1) \textit{Multi-view data:} Our method is trained on multi-view simulation data and we use 4 camera views in our experiments. We test our method without the top-down view because it is similar to the ones in the SDD dataset.
As we see, the performance drops due to the fewer number of data and less diverse views, suggesting that we should use more views in augmentation (which is effortless to do in 3D simulators).

(2) \textit{Random perturbation:} We test our model without random perturbation on the original view trajectory samples by setting $\delta=0$ in Eq.~\eqref{eqn:select_c}.
This leads to the performance drop on all three datasets and particularly on the more difficult Argoverse dataset.

(3) \textit{Adversarial attack:} We test our model without adversarial attack by replacing Eq.~\eqref{eqn:adv} with $V_{1:h}^{adv} = V_{1:h}$.
This is similar to applying the Mixup method~\cite{zhang2017mixup} to two views in the feature space. The performance drops across all three benchmarks. 

(4) \textit{View selection:} We replace Eq.~\eqref{eqn:select_c} with random search to see the effect of view selection. As we see, the significant performance drops, especially on the Stanford Drone dataset, verifying the effectiveness of this design.

\begin{table}[]
\centering
\caption{Performance on ablated versions of our method on three benchmarks. We report 
the minADE\textsubscript{1}($\downarrow$)/minFDE\textsubscript{1}($\downarrow$) metrics.} 
\begin{tabular}{l C{28mm}C{28mm}C{28mm}}
\toprule
Method & VIRAT/ActEV   & Stanford Drone & Argoverse       \\
\midrule
SimAug full model&  21.7 / 42.2 &  15.7 / 30.2 & 67.9 / 175.6\\
\midrule
- top-down view data &  22.8 / 43.6 & 18.4 / 35.6 &  68.4 / 178.3\\ 
- random perturbation & 23.6 / 43.8 & 18.7 / 35.6  & 69.1 / 180.2 \\ 
- adversarial attack &  23.1 / 43.8 & 17.4 / 32.9 & 68.0 / 177.5 \\ 
- view selection &  23.0 / 42.9 & 19.6 / 38.2 & 68.6 / 177.0\\ 
\bottomrule
\end{tabular}
\label{table:ablation}
\end{table}

\section{Conclusion}
\label{sec:concl}
In this paper,
we have introduced \textit{\fancyname}, a novel simulation data augmentation method to learn robust representations for trajectory prediction. Our model is trained only on 3D simulation data and applied out-of-the-box to a wide variety of real video cameras with novel views or scenes. 
We have shown that our method achieves competitive performance on three public benchmarks with and without using the real-world training data.
We believe our approach will facilitate future research and applications on learning robust representation for trajectory prediction with limited or zero training data. Other directions to deal with camera view dependence include using a homography matrix, which may require an additional step of manual or automatic calibration of multiple cameras. We leave them for future work.

\section*{Acknowledgements}
We would like to thank the anonymous reviewers for their useful comments, and Google Cloud for providing GCP research credits. 
This research was supported by NSF grant IIS-1650994, the financial assistance award 60NANB17D156 from NIST, and the Baidu Scholarship. This work was also supported by IARPA via DOI/IBC contract number D17PC00340. The views and conclusions contained herein are those of the authors and should not be interpreted as necessarily representing the official policies or endorsements, either expressed or implied, of IARPA, NIST, DOI/IBC, the National Science Foundation, or the U.S. Government.
\clearpage
%
%
\bibliographystyle{eccv2020/splncs04}
\bibliography{reference}
\end{document}